\documentclass{article}

\usepackage{arxiv}

\usepackage[utf8]{inputenc} 
\usepackage[T1]{fontenc}    
\usepackage{hyperref}       
\usepackage{url}            
\usepackage{booktabs}       
\usepackage{amsfonts}       
\usepackage{nicefrac}       
\usepackage{microtype}      
\usepackage{lipsum}		
\usepackage{graphicx}
\usepackage[numbers]{natbib}
\usepackage{doi}

\usepackage[perpage]{footmisc}

\title{A Survey on Machine Learning Techniques for Auto Labeling of Video, Audio, and Text Data}

\date{} 					

\author{ \href{https://orcid.org/0000-0002-1368-2771}{\includegraphics[scale=0.06]{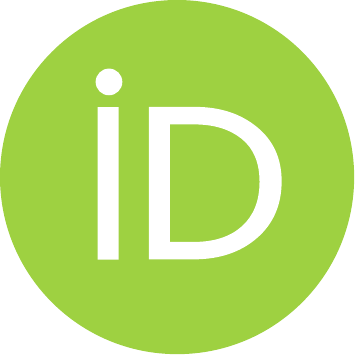}\hspace{1mm}Shikun~Zhang} \\
	Department of Computer Science\\
	New Mexico State University\\
	Las Cruces, NM \\
	\texttt{skzhang@nmsu.edu} \\
	\And
	\href{https://orcid.org/0000-0003-3422-2755}{\includegraphics[scale=0.06]{orcid.pdf}\hspace{1mm}Omid Jafari} \\
	Department of Computer Science\\
	New Mexico State University\\
	Las Cruces, NM \\
	\texttt{ojafari@nmsu.edu} \\
	\And
	\href{https://orcid.org/0000-0001-6284-9251}{\includegraphics[scale=0.06]{orcid.pdf}\hspace{1mm}Parth~Nagarkar} \\
	Department of Computer Science\\
	New Mexico State University\\
	Las Cruces, NM \\
	\texttt{nagarkar@nmsu.edu} \\
}

\begin{document}
\maketitle

\begin{abstract}
	{Machine learning has been utilized to perform tasks in many different domains such as classification, object detection, image segmentation and natural language analysis. Data labeling has always been one of the most important tasks in machine learning. However, labeling large amounts of data increases the monetary cost in machine learning. As a result, researchers started to focus on reducing data annotation and labeling costs. Transfer learning was designed and widely used as an efficient approach that can reasonably reduce the negative impact of limited data, which in turn, reduces the data preparation cost. Even transferring previous knowledge from a source domain reduces the amount of data needed in a target domain. However, large amounts of annotated data are still demanded to build robust models and improve the prediction accuracy of the model. Therefore, researchers started to pay more attention on auto annotation and labeling. In this survey paper, we provide a review of previous techniques that focuses on optimized data annotation and labeling for video, audio, and text data.
}
\end{abstract}


\section{Introduction}
In the machine learning domain a vast amount of labeled data is required. Researchers have gathered datasets containing labeled data from various domains, such as ImageNet \citep{ref6deng2009imagenet}, Lung Image Database Consortium and Image Database Resource Initiative (LIDC-IDRI) \citep{armato2011lung}, MNIST \citep{ref1lecun1998gradient}, etc. Previous work has obtained very exciting results by implementing machine learning techniques with pre-labeled datasets \citep{you2018imagenet, he2019rethinking, ref2tabik2017snapshot, ref3kussul2004improved, ref4schott2018towards, ref5pedamonti2018comparison}. 

To implement machine learning models into a new domain, newly labeled data is needed for training models. However, in some occasions, available data in new tasks are limited. As a result, researchers then studied transfer learning, which is an efficient approach that can reasonably reduce the negative impact of limited data \citep{Ref5pan2009survey, torrey2010transfer, dai2009eigentransfer, houlsby2019parameter, zhang2019computer}. Even with transfer learning that transfers previous knowledge learned from previous tasks to reduce the impact of limited data, annotating data is still a resource consuming task for a real-world implementation with no pre-labeled dataset. Moreover, models also need to be periodically re-trained on new labeled data to keep being robust. Therefore, researchers focus on designing optimized data annotation and labeling frameworks, including automatic and semi-automatic techniques. 

\subsection{Motivation}
{}

\subsubsection{Motivation for Data Annotation and Labeling in Machine Learning}
{Supervised machine learning has achieved widely successes in different domains \citep{kumar2015lung, ramzan2020deep, zhang2019computer, panwar2020application, dong2016multiclass, sprengel2016audio, makantasis2015deep, bharati2016detecting, zheng2019applications, morin2019transmitter}. Data labeling is the most important part of data preparation for supervised learning tasks. Well annotated data are served as input in model training to provide a learning basis for future data processing. Labeled datasets help to train machine learning models to identify and understand the recurring patterns in the input for delivering accurate output. In other words, after being trained on annotated data, machine learning models can be used to recognize the same patterns in the new data that is not seen by the model previously. Since data labeling is an essential task in machine learning domain, lots of well labeled datasets have been already created. 

ImageNet \citep{ref6deng2009imagenet} is one of the most important labeled datasets, which contains more than 14 million images from more than 20,000 categories. Many research have been conducted on ImageNet \citep{ref7krizhevsky2012imagenet, ref8russakovsky2015imagenet, ref9rastegari2016xnor, mishkin2017systematic, you2018imagenet, he2019rethinking, huh2016makes, shankar2020evaluating, he2015delving, xie2020self} and have shown promising results on a such large scale dataset. The amount of time needed to create large scale datasets in a traditional way costs too much. Even for smaller datasets, traditional annotation and labeling will increase the cost of a machine learning task. Therefore, finding methods to optimize the data annotation and labeling process becomes more and more important and a lot of research has been focused on exploring different strategies in reducing data annotation and labeling costs in machine learning tasks.
}

\subsubsection{Motivation of Our Work (Differences from other surveys)}
{The current data annotation and labeling surveys are often focusing on automatic image data annotation. \citep{auty2016survey} conducted a simple survey on image auto-annotation, where it mainly focused on previous work that boosted the performance of image annotation by exploring visual features. Researchers in \citep{zhang2012review} and \citep{bhagat2018image} have reviewed the existing automatic image annotation approaches and they both focus on image retrieval. Similarly, researchers in \citep{zhang2012review} analyzed various automatic image annotation methods, including both feature extraction and semantic learning methods. Researchers in \citep{siddiqui2015survey, wang2011survey} listed some existing automatic image annotation approaches and they both found that existing image auto-annotation methods can be categorized into three major approaches: 1) Generative, 2) Discriminative, and 3) Graph-based. 

\citep{sumathi2011overview} conducted a survey on statistical approaches in automatic image annotation and compared them. \citep{hanbury2008survey} gives an overview of three different image annotation methods: 1) free text annotation, 2) keyword annotation, and 3) annotation using ontologies. \citep{adnan2020survey} has made a detailed study of image annotation methods between manual, semi-automatic, and automatic annotation. Moreover, they talk about the importance of integrating user feedback and a semantic hierarchy into image annotation model. In \citep{cheng2018survey} Nearest neighbor-based image annotation and Tag completion-based image annotation is included into the survey, and the authors have analysed the computational complexity, computation time, and annotation accuracy of different image auto-annotation methods. \citep{tousch2012semantic} is a survey of image annotation uses and user needs, where it also contains comparisons of the annotation approaches using unstructured or hierarchical vocabularies. Moreover, there are other surveys, such as \citep{fu2010survey} and \citep{wang2011active}, that focus on music and multimedia annotation. 

While most existing surveys focus only on image data, in this survey paper, we will provide reviews on optimized annotation and labeling approaches designed for annotating video data, audio data and text data that are based on strategies utilized by researchers. Research that proposes optimized image data annotation and labeling will not be listed in this survey paper since there are many other surveys that have been focusing on image data auto annotation.}

\subsection{Contributions}
{In this paper, we present an in-depth review of previous research that focus on improving the efficiency of data annotation and labeling in the machine learning domain.

Our contributions are listed as following:

\begin{itemize}
    \item We perform an in-depth review over current work in data annotation and labeling and their efficiency improvement strategies. Our review consists of the definitions, the workflow, and the ideas which are proposed in each previous work to improve data annotation efficiency in machine learning tasks.

    \item In video and audio data annotation, we categorize the reviewed previous work based on the strategy that they utilize. For text data, we categorize current works based on specific domains that they focus on.
\end{itemize}
}

\subsection{Paper Organization}
\begin{figure}
	\centering
	\includegraphics[width=\textwidth]{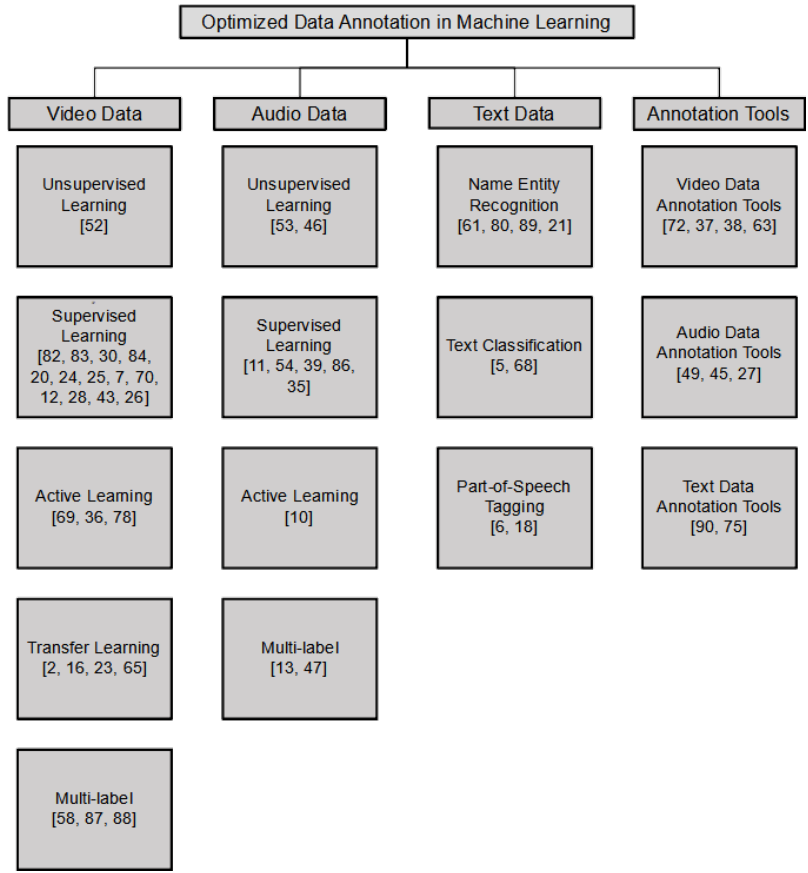}
	\caption{Categorization of Optimized Data Annotation and Labeling Approaches}
	\label{fig:fig1}
\end{figure}

{
The rest of this paper is organized as follows: In section \ref{sec:2}, section \ref{sec:3}, and section \ref{sec:4} we present a detailed review of optimized data annotation techniques designed for video data, audio data, and text data respectively. In section \ref{sec:5}, we talk about existing data annotation tools for video data, audio data, and text data. We then make a conclusion of this paper in section \ref{sec:6}. Finally, we discuss possible future work in section \ref{sec:7}. Moreover, a diagram of this organization is shown in Figure \ref{fig:fig1}.
}

\section{Optimized Annotation for Video Data}
\label{sec:2}
{
A lot of video data is being generated by different data sources every day; hence, a lot of research has focused on video data. In this section, we talk about optimized video annotation techniques that have been proposed so far.

\subsection{Unsupervised Learning Approaches}
{
\citep{moxley2008automatic} proposed an unsupervised automatic video annotation approach that aims to exploit news video to automatically annotate by mining similar videos. The proposed algorithm employs a two-step process of search followed by mining. Given a query video consisting of visual content and speech-recognized transcripts, similar videos are first ranked in a multi-modal search. Later, the transcripts associated with these similar videos are mined to extract keywords for the query. Authors also conducted experiments to show the superiority of the proposed approach.
}

\subsection{Semi-supervised and Supervised Learning Approaches}
{
In \citep{4036892}, exploring the temporal consistency of semantic concepts in video sequences enhances two semi-supervised learning algorithms, which are self-training and co-training. In the enhanced algorithms, the basic sample units took time-constraint shot clusters instead of individual shots. This strategy can obtain more accurate statistical models since most miss-classifications can be corrected before they are applied for re-training. \citep{wang2006automatic} proposes Semi-supervised Learning by Kernel Density Estimation (SSLKDE). This is a novel semi-supervised learning algorithm, and it can avoid the "model assumption" since it is based on a non-parametric method. In the classical Kernel Density Estimation (KDE) approach, only labeled data are utilized while both labeled and unlabeled data are leveraged to estimate class conditional probability densities based on an extended form of KDE in SSLKDE. In \citep{hansen2007automatic}, authors propose an automatic annotation system in which humans pass a surveillance camera. The primary color of the clothing, the height, and focus of attention are associated annotations for each human. The robust background subtraction based on a Codebook representation occurs before annotation. Besides, a body model is used to group similar pixels and estimate the primary colors. A 3D mapping for the head and feet is able to estimate the height while the overall direction of the head shows the focus of attention.

\citep{wang2009unified} proposes optimized multigraph-based semi-supervised learning (OMG-SSL) in order to simultaneously tackle the learning difficulties in a unified scheme. Various crucial factors in video annotation including multiple modalities, multiple distance functions, and temporal consistency can be represented by different graphs since they all correspond to different relationships among video units. Thus, learning with multiple graphs can simultaneously deal with these various crucial factors. Authors point out that this scheme is equivalent to conducting semi-supervised learning on the fused graph after fusing multiple graphs. It can assign suitable weights to the graphs through an optimization approach. They show that a computationally efficient iterative process is able to implement the proposed method. The effectiveness and efficiency of the proposed approach are demonstrated by extensive experiments on the TREC video retrieval evaluation (TRECVID) benchmark. \citep{duchenne2009automatic} uses minimal manual supervision to address the problem of automatic temporal annotation of realistic human actions in video. A kernel-based discriminative clustering algorithm is proposed to locate actions in the weakly-labeled training data. Authors use the obtained motion samples to train the temporal action detectors and apply them to locate the motion in the original video data. Researchers in \citep{gammeter2009know} propose a method to annotate famous tourist sites or buildings. Authors first train their model using large scale crawled data, link the crawled information of the buildings or sights with images, train with images and their linked info. Finally, they run the model on testing images, draw bounding boxes of detected objects, and give related content about the objects within bounding boxes.

In \citep{gao2015optimal}, authors learn an optimal graph (OGL) from multicues (i.e. partial tags and multiple features) and propose a semi-supervised annotation approach. As a result of using their approach, the relationships among the data points can be embedded more accurately. The models are extended to address out-of-sample and noisy label issues. In terms of mean average precision, the consistent superiority of OGL over state-of-the-art methods by up to 12\% which is shown by extensive experiments on four public datasets is promising. In \citep{berg2019semi}, authors state a recursive and semi-automatic annotation approach which proposes initial annotations for all frames in a video based on segmenting only a few manual objects. This novel approach is evaluated on a subset of the RGBT-234 visual-thermal dataset. It uses an advanced video object segmentation method and decreases the workload of a human annotator with approximately 78\%, which outperforms full manual annotation. \citep{sorano2020automatic} describes a method named PassNet to recognize the most frequent events in soccer, such as passes, from video streams. The proposed model combines a set of artificial neural networks that perform feature extraction from video streams detecting object to identify the positions of the ball and the players, and classify frame sequences as passes or not passes. Even when the match video conditions of the test and training sets are considerably different, results show significant improvement in the accuracy of pass detection with respect to baseline classifiers. \citep{chen2020scribblebox} introduces ScribbleBox which is an interactive framework that can annotate object instances with masks in videos. Annotation is divided into two steps. One is annotating objects with tracked boxes, while another is labeling masks inside these tracks. A parametric curve with a small number of control points, which the annotator can interactively correct, is able to be used to approximate the trajectory. This method can efficiently annotate box trackers, and authors show that the J\&F of the ScribbleBox method on DAVIS2017 reached 88.92\%, the average number of clicks per box track was 9.14, and only 4 frames of the video with an average of 65.3 frames needed graffiti annotations.

In \citep{guillermo2020implementation}, authors present an automatic annotation approach by using Mask RCNN in CVAT on an AWS EC2 Instance. Moreover, the viability of deploying data and computing intensive systems on the cloud are completely illustrated. In order to reduce the human labor and time cost while annotating video object bounding box, \citep{le2020toward} proposes Interactive Self-Annotation framework, which is based on recurrent self-supervised learning. The proposed method consists of an Automatic process and an interactive process. The interactive process is able to speed up while a supported detector is built by the automatic process. In the Automatic Recurrent Annotation, an off-the-shelf detector automatically reinforces itself by watching unlabeled videos. This way, better pseudo ground-truth bounding boxes are generated by the trained model of the previous iteration, so that self-supervised training is improved. In the Interactive Recurrent Annotation, the authors use a Hierarchical Correction module. The strength of a Convolutional Neural Network (CNN) for neighbor frames can be used when the annotated frame-distance reduces at each time step. The results on various video datasets show that the human labor and time are reduced effectively by the proposed framework since it can provide high-quality annotations. In \citep{gokalp2021semi}, authors propose a semi-automatic video annotation approach. This approach can eliminate false-positives using temporal information with a tracking-by-detection method through employing multiple hypothesis tracking (MHT). Human operators confirm tracklets which are automatically formed by the MHT method so that the training set can be enlarged. This incremental learning method assists video annotation in an iterative way. Moreover, experiments on AUTH Multidrone Dataset demonstrate that the proposed method can reduce approximately 96\% of the annotation workload.

}
\subsection{Active Learning Approaches}
{

\citep{song2005semi} proposes a novel active learning framework based on multiple complementary predictors and an incremental model adaptation to accelerate the converging speed of the learning process by labeling the most informative samples. \citep{1640557} proposes an active learning framework with clustering tuning. In this framework, an initial training set is first constructed based on clustering the entire video dataset. Then, an SVM-based active learning scheme is proposed to maximize the margin of the SVM classifier by manually and selectively labeling a small set of samples. In each round of active learning, clustering results are tuned based on the prediction results of the current stage. Experimental results show that the proposed framework outperforms the typical active learning algorithms in terms of both annotation accuracy and stability. \citep{vondrick2011video} proposes an adaptive key-frame strategy that utilizes active learning to label only certain objects in certain frames that are helpful in improving the performance. By implementing this proposed method, minimal user effort may achieve high accuracy. Authors demonstrate the framework on four datasets, where two of which are constructed with key frame annotations obtained by Amazon Mechanical Turk. The results indicate that their proposed active learning framework achieves a promising performance.

}

\subsection{Transfer Learning Approaches}

\citep{almajai2012anomaly} aims at constructing an adaptive system for court-based sport video annotation by applying a methodology of anomaly detection. Moreover, authors demonstrate how the detected anomalies can be used to transfer learning from one rule-governed structure to another. \citep{de2013framework} proposes a system that can automatically annotate tennis game videos. A set of mechanisms are embedded in the system to detect anomalies caused by a change of domain in the input data. Once an anomaly is detected, transfer learning methods are there to adapt the knowledge from the source domains to the new domains, such as new sport modalities. Authors also present a generic framework for rule induction that is crucial in adaptive annotation system. In \citep{fuhl2018mam}, authors propose Multiple Annotation Maturation (MAM), which is a new self-training approach. Image data can be fully automatic labeled through MAM and detectors which are produced by MAM can be used afterwards in an online manner. MAM is evaluated based on data from different detection tasks and its performance is compared with the state-of-the-art methods. Results using more than 300,000 images demonstrate promising adaptability and robustness. \citep{shailesh976automatic} focuses on annotating dance videos based on foot postures (stanas). Features from the images are extracted and a deep neural network is used for image classification with the implementation of transfer learning. A trained classifier is used for identifying stanas from the frames of a video which is called Deep Stana Classifier. The proposed annotation system contains Deep Stana Classifier module and an annotation module. The result of the annotations performed on the video is kept in a JSON object format file. Authors mention that their proposed framework is useful for annotating dance videos as well as videos from another domains.

\subsection{Multi-label Annotations Approaches}
{

\citep{qi2007correlative} proposes a novel Correlative Multi-Label (CML) framework which simultaneously classifies concepts and models correlations in a single step. The test results on the TRECVID dataset show the superiority of the proposed framework. \citep{xu2011ensemble} proposes an approach called En-MIMLSVM  for the video annotation task which considers the class imbalance and long time training. Additionally, a temporally consistent weighted multi-instance kernel is developed to consider the temporal consistency in video data as well as the significance of instances of different levels in pyramid representation. The proposed approach shows positive results when evaluated on the  TRECVID 2005 dataset. \citep{xu2012semi} proposes a semi-supervised multi-labels approach that is able to exploit abundant non-annotated videos to help improve the annotation performance. This proposed approach takes label correlations into account and enforces similar instances to share similar multi-labels.
}

}

\section{Optimized Annotation for Audio Data}
\label{sec:3}
{
With the rapidly increased amount of audio data, research have been focusing on performing audio analysis using machine learning techniques. Data annotation and labeling is always a vital important part of these tasks, where it can affect model training results and efficiency. In this section, we talk about papers that focus on optimized audio annotation and labeling. 

\subsection{Unsupervised Learning Approaches}
{
\citep{nam2012learning} presents a sparse feature learning-based data-processing pipeline, which can be implemented in music annotation and retrieval. Music annotation and retrieval systems based on content start with features to process the audio. Experiments show that the newly learned features produce results on the CAL500 dataset using only a linear classifier is comparable with the state-of-the-art music annotation and retrieval systems. \citep{li2020auto} suggests that detecting speech recognition errors and providing possible fixes can provide high quality training data. This auto-annotation system does not need any hand-labeled audios. In the proposed method, an overall word error rate (WER) of the auto-annotated training data is 0.002. Additionally, after applying the auto-suggested fixes, a reduction of 0.907 in WER is obtained.

}

\subsection{Semi-supervised and Supervised Learning Approaches}
{
Labeling any possible sound with great detail could not be accomplished since automatic annotation methods are not mature enough. A taxonomy which represents the world and many classifiers specialized in distinguishing tiny details are required to recognize a general sound. \citep{cano2004automatic} uses WordNet to tackle the taxonomy definition problem. This semantic network can organize real world knowledge. A nearest-neighbor classifier with a database of isolated sounds which is unambiguously linked to WordNet concepts is used to overcome the need for a large number of classifiers to distinguish many different sound classes. A database of more than 50,000 sounds and more than 1,600 concepts achieves 30\% of concept predictions. Moreover, manual annotations is the most significant method for personalized music recommendation systems to query and navigate large music collections. New songs/tracks can only be recommended after manually annotating them. In order to solve this issue, automatic tag annotation based on content analysis is proposed. \citep{ness2009improving} suggests that the performance of the state-of-the-art automatic tag annotation music system which is based on audio content analysis can be improved through stacked generalization. Authors also show the results on two publicly available datasets.

\citep{kojima2016semi} uses semi-automatic annotation to analyze bird songs. Manual annotations are typically required when analyzing the wild recordings. However, the within or between observers may both be varied. Therefore, annotation is not extremely accurate and consistent. Authors propose a system using automated methods from robot audition which is able to accomplish sound source detection, localization, separation, and identification. They also propose to focus on spatial cues and combine other features within a Bayesian framework to do integration instead of studying separately.  Moreover, a large training set of annotated labels are required by a pre-trained model to do supervised machine learning methods. Authors suggest to use less pre-annotation by employing a semi-automatic annotation approach and the experiments of bird songs recordings from the wild show that the proposed system obtains better identification accuracy compared with a method based on conventional robot audition.

The task of personalized feedback is formulated as an effective audio annotation problem by \citep{xu2019affective}. In order to solve the multi-label classification problem, authors propose a novel convolutional clustering neural network (CCNN). They introduce a novel clustering layer to derive intermediate representation instead of aggregating the features of different channels through pooling which can effectively improve annotation performance. They also collect more than 2,000 video clips from the TED website and build an audio annotation dataset to evaluate the performance of their proposed method. Experimental results reveal that the proposed method is better than traditional CNN-based approaches since there is a lower hamming loss for an effective annotation. \citep{ibrahim2020audio} aims at enabling context-aware music recommendation agnostic to user data. In order to explore the relationship between user context and audio content, authors collect track sets using a semi-automatic procedure to help leverage playlist titles as a proxy for context labelling. Consequently, a dataset of 50k tracks labelled with 15 different contexts is created. They next use an audio auto-tagging model to demonstrate benchmark classification results on the created dataset.

}

\subsection{Active Learning Approaches}
{
\citep{broux2016active} aims at helping an annotator to perform a speaker diarization when there are lots of archives in an annotation background. Authors propose a method to correct diarization. The proposed method is mainly evaluated in terms of KSR (Keystroke Saving Rate) and decreases human interventions. As a result, they reduce the number of actions required to correct the speaker diarization output by an absolute value of 6.8

}

\subsection{Multi-label Annotation Approaches}
{
A natural way for searching/annotating music in a large database is query-by-semantic-description (QBSD). \citep{chen2009use} states that using anti-words for each annotation word, which is based on the concept of supervised multiclass labeling (SML), can improve QBSD. Experiments demonstrate that the original SML model achieves only 27.8\% while the annotation system can achieve 31.1\% of equal mean per-word precision and recall when modeling both a word and its anti-word set. Results also show that the retrieval system with anti-word model outperforms, especially when the query keyword has an antonym which is an existing annotation word. \citep{lo2011cost} proposes a new approach which outperforms MIREX 2009 winning method. Authors treat the tag counts as costs so that the audio tagging problem becomes a cost sensitive classification problem. Besides, the audio tagging problem is formulated as a multi-label classification problem by considering the tags co-occurrences. Moreover,  they present that modeling the audio tagging as a novel cost-sensitive multi-label (CSML) learning problem can exploit the tag count and correlation information and give two solutions. Their experiments reveal that the performance is significantly improved compared to MIREX 2009.
}

}

\section{Optimized Annotation for Text Data}
\label{sec:4}
{
The machine learning related research on text data has shown significant success in different domains and researchers have proposed different approaches for annotating text data in an efficient way. In this section, we talk about those papers based on machine learning task that they are used in.

\subsection{Annotation in Named Entity Recognition (NER)}
{
In \citep{rosset2013automatic}, authors point out that human annotation speed and accuracy can be improved by automatic pre-annotation. They also state that the annotators’ subjective assessment cannot always be matched to the actual benefits which are measured in the annotation outcome. In \citep{wang2017employing}, in order to recognize person names in judgment documents, authors propose Aux-LSTM, which is a joint learning method. They utilize various automatic labeling data so that it can help manual labeling data (small size) recognize names. Authors train the auto-annotated data to develop an auxiliary Long Short-Term Memory (LSTM) representation. Moreover, they improve the performance of classifiers which are trained on artificially labeled data by leveraging the auxiliary LSTM representation.

In \citep{yang2018distantly}, authors present a new method in order to handle the distant supervision problems of Chinese NER. According to their method, partial annotation learning is used to reduce the influence of unknown labels of characters. Their method is able to effectively solve the incomplete annotations. They also design an instance selector which is based on reinforcement learning so that positive sentences can be distinguished from auto-generated annotations. It is able to effectively solve the noisy annotations. In their experiments, distant supervision assists them to create two datasets for NER in two domains. Their experimental results demonstrate that their proposed method outperforms the state-of-the-art systems on both two datasets. In \citep{enkhsaikhan2021auto}, authors utilize remote supervision to automatically label dataset in specific domains and point out an iterative deep learning NER framework. They obtain a large BIO-annotated dataset with six geological categories and use this framework on mineral exploration reports. Authors also apply the framework to two other datasets (disease names and chemical names) to verify its generalisation ability. The experiments show that their method works effectively on reducing annotation efforts when identifying a much smaller subset. 

}

\subsection{Annotation in Text Classification}
{
\citep{baruzzo2009general} presents the PIRATES framework which is a Personalized Intelligent Recommender and Annotator TEStbed. It is able to retrieve text-based content and classify them. Users can experiment, customize, and personalize the way they retrieve, filter, and organize information from the Web since this framework utilizes an integrated set of tools. Moreover, the PIRATES framework recommends means of personalized tags and annotate other forms of textual data, which is a new method to automate typical manual tasks including content annotation and tagging. In \citep{song2019employing}, authors propose an information retrieval approach for government documents. This approach is able to build a large-scale automatic annotated dataset. The experimental results demonstrate that the supervised classification model which is trained on automatic constructed dataset shows better results compared with the baseline method.
}

\subsection{Annotation in Part-of-speech Tagging}
{
\citep{beisswenger2016empirist} describes a task which is automatic linguistic annotation of German language data. The data come from genres of computer-mediated communication (CMC), social media interactions, and Web corpora. Authors perform tokenization and part-of-speech tagging on two data sets. Their best tokenizer achieves an F1- score of 99.57\% (compared to the 98.95\% off-the-shelf baseline). Moreover, their best tagger achieves an accuracy of 90.44\% (compared to the 84.86\% baseline). The task results show a significant improvement compared with current off-the-shelf tools for German. Additionally, in order to annotate languages without natural definitions of words, an annotation system with feasibility and flexibility for joint tokenization and part-of-speech (POS) tagging is needed. As a result, \citep{ding2018nova} proposes nova, an annotation system which contains only four basic tags (n, v, a, and o). The tags are able to be further modified and combined so that they can adapt complex linguistic phenomena in tokenization and POS tagging. Authors also discuss the relation between nova and two universal POS tagsets.
}

\section{Optimized Data Annotation Tools}
\label{sec:5}

In the previous sections, we listed papers focusing on optimized data annotation approaches. In this section, we give a review of the existing data annotation tools that implement automatic or semi-automatic data annotation approaches.

\subsection{Annotation Tools for Video Data}
{
\citep{9031469} develops a web tool, which can be used to annotate underwater videos. Users can also create a short tracking video for each annotation that shows how an annotated concept moves in time. Verifying the accuracy of existing annotations is supported by the tool. Additionally, users may create a neural network model from existing annotations and automatically annotate new videos using previously created models. \citep{kim2013semi} develops a semi-automatic video annotation tool that can automatically generate the initial annotation data for input videos by the automatic object detection modules. The system also has several user-friendly functions to allow the users to check the validity of the initial annotation data. Using this developed video annotation tool, users can generate large amount of ground truth data for videos. \citep{kithmi2016semantic} proposes a tool for tagging and annotation. The tool can analyze the video structure to detect shot boundaries where shots in each video are identified using image duplication techniques. A single frame from each shot is passed to a deep learning model implemented using TensorFlow, which is trained for feature extraction and classification of objects in each frame. Moreover, an automatic textual annotation is generated for each video. \citep{sanchez2020semi} proposes an novel tool named Annotation as a Service (AaaS) that is designed to integrate heterogeneous video annotation workflows into containers and take advantage of a cloud service that is highly scalable and reliable based on Kubernetes workloads. The solution has proven to be efficient and resilient and automatic pre-annotations with the proposed strategy reduce the time of human participation in the annotation by up to 80\% maximum and 60\% on average.

}

\subsection{Annotation Tools for Audio Data}
{
\citep{marques1999automatic} describes a system which can automatically annotate audio files. The application outputs time-aligned text labels describing the files audio content when there is a sound file as input. Labels of eight musical instrument names and the label ’other’ are both included in this annotation system. Two sound classifiers, which are built after experimenting using different parameters, are applied by the annotation tool. The first classifier uses Gaussian Mixture Models and the mel cepstral feature set so that instrument set and non instrument set sounds can be divided. It needs 0.2 seconds to correctly classify an audio segment with 75\% accuracy. The second classifier uses Support Vector Machines and mel cepstral feature set to distinguish the type of instruments. The time length is 0.2 seconds, while the accuracy is 70\%. In \citep{levy2019gecko}, authors propose a tool that can annotate speech and language features of conversations, named Gecko. Gecko is able to segment the voice signal from the speaker and annotate the linguistic content of the conversation efficiently and effectively. Authors also state that the presentation of the automatic segmentation output as well as transcription systems in an intuitive user interface for editing is Gecko’s significant feature. Voice Activity Detection (VAD), Diarization, Speaker Identification, and ASR outputs are allowed to be annotated on a large scale through Gecko. Gecko is publicly available for the benefit of the community \footnote{https://github.com/gong-io/gecko}. In \citep{grover2020audino}, authors propose audino, which is a tool to annotate audio and speech in a collaborative and modern way. Temporal segmentation in audios can be defined and described by annotators and a dynamically generated form can label and transcribe the segments. Moreover, user roles and project assignments are able to be controlled by the admin dashboard which describes labels and values. Furthermore, JSON format is available when exporting annotations for further processing. A key-based API can assist user upload and assign audio data in the tool. Authors also emphasize on the flexibility of this annotation tool.

}

\subsection{Annotation Tools for Text Data}
{
In order to manually annotate text documents, \citep{yimam2014automatic} presents a flexible approach which is extending an open-source web-based annotation tool named WebAnno. An arbitrary number of layers, which do separate or simultaneous annotation and support most types of linguistic annotations, are possible to be added and configured through a web-based UI. A generic machine learning component is tightly integrated to automatically annotate suggestions of span annotations. Two studies demonstrate that automatic annotation suggestions which are combined with the split-pane UI concept reduces annotation time effectively. Additionally, \citep{taycsimobile} presents a mobile annotation tool featuring user-friendly and flexibility. The proposed tool provides auto-annotation, label management, and multi-user support.
}

}

\section{Conclusion}
\label{sec:6}
{
Data annotation is an important task of machine learning. Optimized data annotation including automatic and semi-automatic approaches increases the data annotation efficiency and reduces the cost of machine learning tasks. In this paper, we categorized the optimized data annotation approaches designed for video data, audio data, and text data. For video and audio data, we categorized optimized annotation techniques into unsupervised learning approaches, semi-supervised learning approaches, supervised learning approaches, active learning approaches, and approaches that utilize transfer learning. For text data, we categorized existing works by the domains that they are focusing on. Finally, we discussed about existing annotation tools.
}

\section{Future Work}
\label{sec:7}
{
Previous research have proposed many efficient approaches to optimize the data annotation in machine learning including the ones that focus on unsupervised, semi-supervised, and supervised learning, active learning, and transfer learning approaches. Although data annotations have been fully researched, further endeavor can still be made by designing a transfer learning based active learning framework, which is an active learning framework that transfer previous knowledge to further accelerate the data annotation tasks. By combining active learning and transfer learning, efficiency may be further improved. Future research can also explore possible human-in-the-loop active learning frameworks, which may result in higher accuracy with minimal human efforts involved. Deep reinforcement learning should also be paid more concentration on as it can be implemented to automatically learn and improve data annotation based on rewards, which can provide a more efficient and automatic approach for some real-world implementation scenarios. Another possible future direction is heterogeneous data. Nowadays, a lot of surveillance devices are being used to monitor a specific location. These devices can generate different types of data (e.g. temperature and humidity received from sensors, video and image received from cameras, and audio received from microphones). Since all of these heterogeneous data can belong to the same location, labeling them is a challenging research topic.
}

\bibliographystyle{plainnat}
\bibliography{references}  

\begin{thebibliography}{94}
\providecommand{\natexlab}[1]{#1}
\providecommand{\url}[1]{\texttt{#1}}
\expandafter\ifx\csname urlstyle\endcsname\relax
  \providecommand{\doi}[1]{doi: #1}\else
  \providecommand{\doi}{doi: \begingroup \urlstyle{rm}\Url}\fi

\bibitem[Adnan et~al.(2020)Adnan, Rahim, Al-Jawaheri, Ali, Waheed, and
  Radie]{adnan2020survey}
Myasar~Mundher Adnan, Mohd Shafry~Mohd Rahim, Kerrar Al-Jawaheri,
  Mohammed~Hasan Ali, Safa~Riyadh Waheed, and A~Hussien Radie.
\newblock A survey and analysis on image annotation.
\newblock In \emph{2020 3rd International Conference on Engineering Technology
  and its Applications (IICETA)}, pages 203--208. IEEE, 2020.

\bibitem[Almajai et~al.(2012)Almajai, Yan, de~Campos, Khan, Christmas,
  Windridge, and Kittler]{almajai2012anomaly}
Ibrahim Almajai, Fei Yan, Teofilo de~Campos, Aftab Khan, William Christmas,
  David Windridge, and Josef Kittler.
\newblock Anomaly detection and knowledge transfer in automatic sports video
  annotation.
\newblock In \emph{Detection and identification of rare audiovisual cues},
  pages 109--117. Springer, 2012.

\bibitem[Armato~III et~al.(2011)Armato~III, McLennan, Bidaut, McNitt-Gray,
  Meyer, Reeves, Zhao, Aberle, Henschke, Hoffman, et~al.]{armato2011lung}
Samuel~G Armato~III, Geoffrey McLennan, Luc Bidaut, Michael~F McNitt-Gray,
  Charles~R Meyer, Anthony~P Reeves, Binsheng Zhao, Denise~R Aberle, Claudia~I
  Henschke, Eric~A Hoffman, et~al.
\newblock The lung image database consortium (lidc) and image database resource
  initiative (idri): a completed reference database of lung nodules on ct
  scans.
\newblock \emph{Medical physics}, 38\penalty0 (2):\penalty0 915--931, 2011.

\bibitem[Auty and Phakatkar(2016)]{auty2016survey}
Smita Auty and Anupama Phakatkar.
\newblock A survey on: Image auto-annotation by visual features.
\newblock 2016.

\bibitem[Baruzzo et~al.(2009)Baruzzo, Dattolo, Pudota, and
  Tasso]{baruzzo2009general}
Andrea Baruzzo, Antonina Dattolo, Nirmala Pudota, and Carlo Tasso.
\newblock A general framework for personalized text classification and
  annotation.
\newblock In \emph{International Workshop on Adaptation and Personalization for
  Web}, volume~2, pages 31--39. Citeseer, 2009.

\bibitem[Bei{\ss}wenger et~al.(2016)Bei{\ss}wenger, Bartsch, Evert, and
  W{\"u}rzner]{beisswenger2016empirist}
Michael Bei{\ss}wenger, Sabine Bartsch, Stefan Evert, and Kay-Michael
  W{\"u}rzner.
\newblock Empirist 2015: A shared task on the automatic linguistic annotation
  of computer-mediated communication and web corpora.
\newblock In \emph{Proceedings of the 10th Web as Corpus Workshop}, pages
  44--56, 2016.

\bibitem[Berg et~al.(2019)Berg, Johnander, Durand~de Gevigney, Ahlberg, and
  Felsberg]{berg2019semi}
Amanda Berg, Joakim Johnander, Flavie Durand~de Gevigney, Jorgen Ahlberg, and
  Michael Felsberg.
\newblock Semi-automatic annotation of objects in visual-thermal video.
\newblock In \emph{Proceedings of the IEEE/CVF International Conference on
  Computer Vision Workshops}, pages 0--0, 2019.

\bibitem[Bhagat and Choudhary(2018)]{bhagat2018image}
PK~Bhagat and Prakash Choudhary.
\newblock Image annotation: Then and now.
\newblock \emph{Image and Vision Computing}, 80:\penalty0 1--23, 2018.

\bibitem[Bharati et~al.(2016)Bharati, Singh, Vatsa, and
  Bowyer]{bharati2016detecting}
Aparna Bharati, Richa Singh, Mayank Vatsa, and Kevin~W Bowyer.
\newblock Detecting facial retouching using supervised deep learning.
\newblock \emph{IEEE Transactions on Information Forensics and Security},
  11\penalty0 (9):\penalty0 1903--1913, 2016.

\bibitem[Broux et~al.(2016)Broux, Doukhan, Petitrenaud, Meignier, and
  Carrive]{broux2016active}
Pierre-Alexandre Broux, David Doukhan, Simon Petitrenaud, Sylvain Meignier, and
  Jean Carrive.
\newblock An active learning method for speaker identity annotation in audio
  recordings.
\newblock In \emph{1st International Workshop on Multimodal Media Data
  Analytics (MMDA 2016)}, 2016.

\bibitem[Cano and Koppenberger(2004)]{cano2004automatic}
Pedro Cano and Markus Koppenberger.
\newblock Automatic sound annotation.
\newblock In \emph{Proceedings of the 2004 14th IEEE Signal Processing Society
  Workshop Machine Learning for Signal Processing, 2004.}, pages 391--400.
  IEEE, 2004.

\bibitem[Chen et~al.(2020)Chen, Ling, Zeng, Gao, Xu, and
  Fidler]{chen2020scribblebox}
Bowen Chen, Huan Ling, Xiaohui Zeng, Jun Gao, Ziyue Xu, and Sanja Fidler.
\newblock Scribblebox: Interactive annotation framework for video object
  segmentation.
\newblock In \emph{Computer Vision--ECCV 2020: 16th European Conference,
  Glasgow, UK, August 23--28, 2020, Proceedings, Part XIII 16}, pages 293--310.
  Springer, 2020.

\bibitem[Chen and Jang(2009)]{chen2009use}
Zhi-Sheng Chen and Jyh-Shing~Roger Jang.
\newblock On the use of anti-word models for audio music annotation and
  retrieval.
\newblock \emph{IEEE transactions on audio, speech, and language processing},
  17\penalty0 (8):\penalty0 1547--1556, 2009.

\bibitem[Cheng et~al.(2018)Cheng, Zhang, Fu, Tu, and Li]{cheng2018survey}
Qimin Cheng, Qian Zhang, Peng Fu, Conghuan Tu, and Sen Li.
\newblock A survey and analysis on automatic image annotation.
\newblock \emph{Pattern Recognition}, 79:\penalty0 242--259, 2018.

\bibitem[Dai et~al.(2009)Dai, Jin, Xue, Yang, and Yu]{dai2009eigentransfer}
Wenyuan Dai, Ou~Jin, Gui-Rong Xue, Qiang Yang, and Yong Yu.
\newblock Eigentransfer: a unified framework for transfer learning.
\newblock In \emph{Proceedings of the 26th Annual International Conference on
  Machine Learning}, pages 193--200, 2009.

\bibitem[De~Campos et~al.(2013)De~Campos, Khan, Yan, FarajiDavar, Windridge,
  Kittler, and Christmas]{de2013framework}
TE~De~Campos, Aftab Khan, Fei Yan, Nazli FarajiDavar, David Windridge, Josef
  Kittler, and William Christmas.
\newblock A framework for automatic sports video annotation with anomaly
  detection and transfer learning.
\newblock \emph{Machine Learning and Cognitive Science, collocated with
  EUCOGIII}, 2013.

\bibitem[Deng et~al.(2009)Deng, Dong, Socher, Li, Li, and
  Fei-Fei]{ref6deng2009imagenet}
Jia Deng, Wei Dong, Richard Socher, Li-Jia Li, Kai Li, and Li~Fei-Fei.
\newblock Imagenet: A large-scale hierarchical image database.
\newblock In \emph{2009 IEEE conference on computer vision and pattern
  recognition}, pages 248--255. Ieee, 2009.

\bibitem[Ding et~al.(2018)Ding, Utiyama, and Sumita]{ding2018nova}
Chenchen Ding, Masao Utiyama, and Eiichiro Sumita.
\newblock Nova: A feasible and flexible annotation system for joint
  tokenization and part-of-speech tagging.
\newblock \emph{ACM Transactions on Asian and Low-Resource Language Information
  Processing (TALLIP)}, 18\penalty0 (2):\penalty0 1--18, 2018.

\bibitem[Dong et~al.(2016)Dong, Qian, Guan, Huang, Yu, and
  Yang]{dong2016multiclass}
Xishuang Dong, Lijun Qian, Yi~Guan, Lei Huang, Qiubin Yu, and Jinfeng Yang.
\newblock A multiclass classification method based on deep learning for named
  entity recognition in electronic medical records.
\newblock In \emph{2016 New York Scientific Data Summit (NYSDS)}, pages 1--10.
  IEEE, 2016.

\bibitem[Duchenne et~al.(2009)Duchenne, Laptev, Sivic, Bach, and
  Ponce]{duchenne2009automatic}
Olivier Duchenne, Ivan Laptev, Josef Sivic, Francis Bach, and Jean Ponce.
\newblock Automatic annotation of human actions in video.
\newblock In \emph{2009 IEEE 12th International Conference on Computer Vision},
  pages 1491--1498. IEEE, 2009.

\bibitem[Enkhsaikhan et~al.(2021)Enkhsaikhan, Liu, Holden, and
  Duuring]{enkhsaikhan2021auto}
Majigsuren Enkhsaikhan, Wei Liu, Eun-Jung Holden, and Paul Duuring.
\newblock Auto-labelling entities in low-resource text: a geological case
  study.
\newblock \emph{Knowledge and Information Systems}, 63\penalty0 (3):\penalty0
  695--715, 2021.

\bibitem[Fu et~al.(2010)Fu, Lu, Ting, and Zhang]{fu2010survey}
Zhouyu Fu, Guojun Lu, Kai~Ming Ting, and Dengsheng Zhang.
\newblock A survey of audio-based music classification and annotation.
\newblock \emph{IEEE transactions on multimedia}, 13\penalty0 (2):\penalty0
  303--319, 2010.

\bibitem[Fuhl et~al.(2018)Fuhl, Castner, Zhuang, Holzer, Rosenstiel, and
  Kasneci]{fuhl2018mam}
Wolfgang Fuhl, Nora Castner, Lin Zhuang, Markus Holzer, Wolfgang Rosenstiel,
  and Enkelejda Kasneci.
\newblock Mam: Transfer learning for fully automatic video annotation and
  specialized detector creation.
\newblock In \emph{Proceedings of the European Conference on Computer Vision
  (ECCV) Workshops}, pages 0--0, 2018.

\bibitem[Gammeter et~al.(2009)Gammeter, Bossard, Quack, and
  Van~Gool]{gammeter2009know}
Stephan Gammeter, Lukas Bossard, Till Quack, and Luc Van~Gool.
\newblock I know what you did last summer: object-level auto-annotation of
  holiday snaps.
\newblock In \emph{2009 IEEE 12th International Conference on Computer Vision},
  pages 614--621. IEEE, 2009.

\bibitem[Gao et~al.(2015)Gao, Song, Nie, Yan, Sebe, and
  Tao~Shen]{gao2015optimal}
Lianli Gao, Jingkuan Song, Feiping Nie, Yan Yan, Nicu Sebe, and Heng Tao~Shen.
\newblock Optimal graph learning with partial tags and multiple features for
  image and video annotation.
\newblock In \emph{Proceedings of the IEEE Conference on Computer Vision and
  Pattern Recognition}, pages 4371--4379, 2015.

\bibitem[Gokalp~Ince et~al.(2021)Gokalp~Ince, Koksal, Fazla, and
  Aydin~Alatan]{gokalp2021semi}
Kutalmis Gokalp~Ince, Aybora Koksal, Arda Fazla, and A~Aydin~Alatan.
\newblock Semi-automatic video annotation for object detection.
\newblock \emph{arXiv e-prints}, pages arXiv--2101, 2021.

\bibitem[Grover et~al.(2020)Grover, Bamdev, Kumar, Hama, and
  Shah]{grover2020audino}
Manraj~Singh Grover, Pakhi Bamdev, Yaman Kumar, Mika Hama, and Rajiv~Ratn Shah.
\newblock audino: A modern annotation tool for audio and speech.
\newblock \emph{arXiv preprint arXiv:2006.05236}, 2020.

\bibitem[Guillermo et~al.(2020)Guillermo, Billones, Bandala, Vicerra, Sybingco,
  Dadios, and Fillone]{guillermo2020implementation}
Marielet Guillermo, Robert~Kerwin Billones, Argel Bandala, Ryan~Rhay Vicerra,
  Edwin Sybingco, Elmer~P Dadios, and Alexis Fillone.
\newblock Implementation of automated annotation through mask rcnn object
  detection model in cvat using aws ec2 instance.
\newblock In \emph{2020 IEEE region 10 conference (TENCON)}, pages 708--713.
  IEEE, 2020.

\bibitem[Hanbury(2008)]{hanbury2008survey}
Allan Hanbury.
\newblock A survey of methods for image annotation.
\newblock \emph{Journal of Visual Languages \& Computing}, 19\penalty0
  (5):\penalty0 617--627, 2008.

\bibitem[Hansen et~al.(2007)Hansen, Mortensen, Duizer, Andersen, and
  Moeslund]{hansen2007automatic}
Dennis~M{\o}lholm Hansen, Bjarne~K Mortensen, PT~Duizer, Jens~R Andersen, and
  Thomas~B Moeslund.
\newblock Automatic annotation of humans in surveillance video.
\newblock In \emph{Fourth Canadian Conference on Computer and Robot Vision
  (CRV'07)}, pages 473--480. IEEE, 2007.

\bibitem[He et~al.(2015)He, Zhang, Ren, and Sun]{he2015delving}
Kaiming He, Xiangyu Zhang, Shaoqing Ren, and Jian Sun.
\newblock Delving deep into rectifiers: Surpassing human-level performance on
  imagenet classification.
\newblock In \emph{Proceedings of the IEEE international conference on computer
  vision}, pages 1026--1034, 2015.

\bibitem[He et~al.(2019)He, Girshick, and Doll{\'a}r]{he2019rethinking}
Kaiming He, Ross Girshick, and Piotr Doll{\'a}r.
\newblock Rethinking imagenet pre-training.
\newblock In \emph{Proceedings of the IEEE/CVF International Conference on
  Computer Vision}, pages 4918--4927, 2019.

\bibitem[Houlsby et~al.(2019)Houlsby, Giurgiu, Jastrzebski, Morrone,
  De~Laroussilhe, Gesmundo, Attariyan, and Gelly]{houlsby2019parameter}
Neil Houlsby, Andrei Giurgiu, Stanislaw Jastrzebski, Bruna Morrone, Quentin
  De~Laroussilhe, Andrea Gesmundo, Mona Attariyan, and Sylvain Gelly.
\newblock Parameter-efficient transfer learning for nlp.
\newblock In \emph{International Conference on Machine Learning}, pages
  2790--2799. PMLR, 2019.

\bibitem[Huh et~al.(2016)Huh, Agrawal, and Efros]{huh2016makes}
Minyoung Huh, Pulkit Agrawal, and Alexei~A Efros.
\newblock What makes imagenet good for transfer learning?
\newblock \emph{arXiv preprint arXiv:1608.08614}, 2016.

\bibitem[Ibrahim et~al.(2020)Ibrahim, Royo-Letelier, Epure, Peeters, and
  Richard]{ibrahim2020audio}
Karim~M Ibrahim, Jimena Royo-Letelier, Elena~V Epure, Geoffroy Peeters, and
  Gael Richard.
\newblock Audio-based auto-tagging with contextual tags for music.
\newblock In \emph{ICASSP 2020-2020 IEEE International Conference on Acoustics,
  Speech and Signal Processing (ICASSP)}, pages 16--20. IEEE, 2020.

\bibitem[jun Qi et~al.(2006)jun Qi, Song, Hua, Zhang, and Dai]{1640557}
Guo jun Qi, Yan Song, Xian-Sheng Hua, Hong-Jiang Zhang, and Li-Rong Dai.
\newblock Video annotation by active learning and cluster tuning.
\newblock In \emph{2006 Conference on Computer Vision and Pattern Recognition
  Workshop (CVPRW'06)}, pages 114--114, 2006.
\newblock \doi{10.1109/CVPRW.2006.211}.

\bibitem[Kim et~al.(2013)Kim, Gwon, Park, Kim, and Kim]{kim2013semi}
Joosung Kim, Ryu-Hyeok Gwon, Jin-Tak Park, Hakil Kim, and Yoo-Sung Kim.
\newblock A semi-automatic video annotation tool to generate ground truth for
  intelligent video surveillance systems.
\newblock In \emph{Proceedings of International Conference on Advances in
  Mobile Computing \& Multimedia}, pages 509--513, 2013.

\bibitem[Kithmi(2016)]{kithmi2016semantic}
Ashangani Kithmi.
\newblock Semantic video search by automatic video annotation using tensorflow.
\newblock In \emph{Manufacturing \& Industrial Engineering Symposium (MIES).},
  2016.

\bibitem[Kojima et~al.(2016)Kojima, Sugiyama, Suzuki, Nakadai, and
  Taylor]{kojima2016semi}
Ryosuke Kojima, Osamu Sugiyama, Reiji Suzuki, Kazuhiro Nakadai, and Charles~E
  Taylor.
\newblock Semi-automatic bird song analysis by spatial-cue-based integration of
  sound source detection, localization, separation, and identification.
\newblock In \emph{2016 IEEE/RSJ International Conference on Intelligent Robots
  and Systems (IROS)}, pages 1287--1292. IEEE, 2016.

\bibitem[Krizhevsky et~al.(2012)Krizhevsky, Sutskever, and
  Hinton]{ref7krizhevsky2012imagenet}
Alex Krizhevsky, Ilya Sutskever, and Geoffrey~E Hinton.
\newblock Imagenet classification with deep convolutional neural networks.
\newblock \emph{Advances in neural information processing systems},
  25:\penalty0 1097--1105, 2012.

\bibitem[Kumar et~al.(2015)Kumar, Wong, and Clausi]{kumar2015lung}
Devinder Kumar, Alexander Wong, and David~A Clausi.
\newblock Lung nodule classification using deep features in ct images.
\newblock In \emph{2015 12th Conference on Computer and Robot Vision}, pages
  133--138. IEEE, 2015.

\bibitem[Kussul and Baidyk(2004)]{ref3kussul2004improved}
Ernst Kussul and Tatiana Baidyk.
\newblock Improved method of handwritten digit recognition tested on mnist
  database.
\newblock \emph{Image and Vision Computing}, 22\penalty0 (12):\penalty0
  971--981, 2004.

\bibitem[Le et~al.(2020)Le, Sugimoto, Ono, and Kawasaki]{le2020toward}
Trung-Nghia Le, Akihiro Sugimoto, Shintaro Ono, and Hiroshi Kawasaki.
\newblock Toward interactive self-annotation for video object bounding box:
  Recurrent self-learning and hierarchical annotation based framework.
\newblock In \emph{Proceedings of the IEEE/CVF Winter Conference on
  Applications of Computer Vision}, pages 3231--3240, 2020.

\bibitem[LeCun et~al.(1998)LeCun, Bottou, Bengio, and
  Haffner]{ref1lecun1998gradient}
Yann LeCun, L{\'e}on Bottou, Yoshua Bengio, and Patrick Haffner.
\newblock Gradient-based learning applied to document recognition.
\newblock \emph{Proceedings of the IEEE}, 86\penalty0 (11):\penalty0
  2278--2324, 1998.

\bibitem[Levy et~al.(2019)Levy, Sitman, Amir, Golshtein, Mochary, Reshef,
  Reichart, and Allouche]{levy2019gecko}
Golan Levy, Raquel Sitman, Ido Amir, Eduard Golshtein, Ran Mochary, Eilon
  Reshef, Roi Reichart, and Omri Allouche.
\newblock Gecko-a tool for effective annotation of human conversations.
\newblock In \emph{INTERSPEECH}, pages 3677--3678, 2019.

\bibitem[Li and Ture(2020)]{li2020auto}
Wenyan Li and Ferhan Ture.
\newblock Auto-annotation for voice-enabled entertainment systems.
\newblock In \emph{Proceedings of the 43rd International ACM SIGIR Conference
  on Research and Development in Information Retrieval}, pages 1557--1560,
  2020.

\bibitem[Lo et~al.(2011)Lo, Wang, Wang, and Lin]{lo2011cost}
Hung-Yi Lo, Ju-Chiang Wang, Hsin-Min Wang, and Shou-De Lin.
\newblock Cost-sensitive multi-label learning for audio tag annotation and
  retrieval.
\newblock \emph{IEEE Transactions on Multimedia}, 13\penalty0 (3):\penalty0
  518--529, 2011.

\bibitem[Makantasis et~al.(2015)Makantasis, Karantzalos, Doulamis, and
  Doulamis]{makantasis2015deep}
Konstantinos Makantasis, Konstantinos Karantzalos, Anastasios Doulamis, and
  Nikolaos Doulamis.
\newblock Deep supervised learning for hyperspectral data classification
  through convolutional neural networks.
\newblock In \emph{2015 IEEE International Geoscience and Remote Sensing
  Symposium (IGARSS)}, pages 4959--4962. IEEE, 2015.

\bibitem[Marques(1999)]{marques1999automatic}
Janet Marques.
\newblock \emph{An automatic annotation system for audio data containing
  music}.
\newblock PhD thesis, Massachusetts Institute of Technology, 1999.

\bibitem[Mishkin et~al.(2017)Mishkin, Sergievskiy, and
  Matas]{mishkin2017systematic}
Dmytro Mishkin, Nikolay Sergievskiy, and Jiri Matas.
\newblock Systematic evaluation of convolution neural network advances on the
  imagenet.
\newblock \emph{Computer Vision and Image Understanding}, 161:\penalty0 11--19,
  2017.

\bibitem[Morin et~al.(2019)Morin, Cardoso, Hoydis, Gorce, and
  Vial]{morin2019transmitter}
Cyrille Morin, Leonardo~S Cardoso, Jakob Hoydis, Jean-Marie Gorce, and Thibaud
  Vial.
\newblock Transmitter classification with supervised deep learning.
\newblock In \emph{International Conference on Cognitive Radio Oriented
  Wireless Networks}, pages 73--86. Springer, 2019.

\bibitem[Moxley et~al.(2008)Moxley, Mei, Hua, Ma, and
  Manjunath]{moxley2008automatic}
Emily Moxley, Tao Mei, Xian-Sheng Hua, Wei-Ying Ma, and BS~Manjunath.
\newblock Automatic video annotation through search and mining.
\newblock In \emph{2008 IEEE international conference on multimedia and expo},
  pages 685--688. IEEE, 2008.

\bibitem[Nam et~al.(2012)Nam, Herrera, Slaney, and Smith~III]{nam2012learning}
Juhan Nam, Jorge Herrera, Malcolm Slaney, and Julius~O Smith~III.
\newblock Learning sparse feature representations for music annotation and
  retrieval.
\newblock In \emph{ISMIR}, pages 565--570. Citeseer, 2012.

\bibitem[Ness et~al.(2009)Ness, Theocharis, Tzanetakis, and
  Martins]{ness2009improving}
Steven~R Ness, Anthony Theocharis, George Tzanetakis, and Luis~Gustavo Martins.
\newblock Improving automatic music tag annotation using stacked generalization
  of probabilistic svm outputs.
\newblock In \emph{Proceedings of the 17th ACM international conference on
  Multimedia}, pages 705--708, 2009.

\bibitem[Pan and Yang(2009)]{Ref5pan2009survey}
Sinno~Jialin Pan and Qiang Yang.
\newblock A survey on transfer learning.
\newblock \emph{IEEE Transactions on knowledge and data engineering},
  22\penalty0 (10):\penalty0 1345--1359, 2009.

\bibitem[Panwar et~al.(2020)Panwar, Gupta, Siddiqui, Morales-Menendez, and
  Singh]{panwar2020application}
Harsh Panwar, PK~Gupta, Mohammad~Khubeb Siddiqui, Ruben Morales-Menendez, and
  Vaishnavi Singh.
\newblock Application of deep learning for fast detection of covid-19 in x-rays
  using ncovnet.
\newblock \emph{Chaos, Solitons \& Fractals}, 138:\penalty0 109944, 2020.

\bibitem[Pedamonti(2018)]{ref5pedamonti2018comparison}
Dabal Pedamonti.
\newblock Comparison of non-linear activation functions for deep neural
  networks on mnist classification task.
\newblock \emph{arXiv preprint arXiv:1804.02763}, 2018.

\bibitem[Qi et~al.(2007)Qi, Hua, Rui, Tang, Mei, and Zhang]{qi2007correlative}
Guo-Jun Qi, Xian-Sheng Hua, Yong Rui, Jinhui Tang, Tao Mei, and Hong-Jiang
  Zhang.
\newblock Correlative multi-label video annotation.
\newblock In \emph{Proceedings of the 15th ACM international conference on
  Multimedia}, pages 17--26, 2007.

\bibitem[Ramzan et~al.(2020)Ramzan, Khan, Rehmat, Iqbal, Saba, Rehman, and
  Mehmood]{ramzan2020deep}
Farheen Ramzan, Muhammad Usman~Ghani Khan, Asim Rehmat, Sajid Iqbal, Tanzila
  Saba, Amjad Rehman, and Zahid Mehmood.
\newblock A deep learning approach for automated diagnosis and multi-class
  classification of alzheimer’s disease stages using resting-state fmri and
  residual neural networks.
\newblock \emph{Journal of medical systems}, 44\penalty0 (2):\penalty0 1--16,
  2020.

\bibitem[Rastegari et~al.(2016)Rastegari, Ordonez, Redmon, and
  Farhadi]{ref9rastegari2016xnor}
Mohammad Rastegari, Vicente Ordonez, Joseph Redmon, and Ali Farhadi.
\newblock Xnor-net: Imagenet classification using binary convolutional neural
  networks.
\newblock In \emph{European conference on computer vision}, pages 525--542.
  Springer, 2016.

\bibitem[Rosset et~al.(2013)Rosset, Grouin, Lavergne, Jannet, Leixa, Galibert,
  and Zweigenbaum]{rosset2013automatic}
Sophie Rosset, Cyril Grouin, Thomas Lavergne, Mohamed~Ben Jannet,
  J{\'e}r{\'e}my Leixa, Olivier Galibert, and Pierre Zweigenbaum.
\newblock Automatic named entity pre-annotation for out-of-domain human
  annotation.
\newblock In \emph{Proceedings of the 7th Linguistic Annotation Workshop and
  Interoperability with Discourse}, pages 168--177, 2013.

\bibitem[Russakovsky et~al.(2015)Russakovsky, Deng, Su, Krause, Satheesh, Ma,
  Huang, Karpathy, Khosla, Bernstein, et~al.]{ref8russakovsky2015imagenet}
Olga Russakovsky, Jia Deng, Hao Su, Jonathan Krause, Sanjeev Satheesh, Sean Ma,
  Zhiheng Huang, Andrej Karpathy, Aditya Khosla, Michael Bernstein, et~al.
\newblock Imagenet large scale visual recognition challenge.
\newblock \emph{International journal of computer vision}, 115\penalty0
  (3):\penalty0 211--252, 2015.

\bibitem[S{\'a}nchez-Carballido et~al.(2020)S{\'a}nchez-Carballido, Senderos,
  Nieto, and Otaegui]{sanchez2020semi}
Sergio S{\'a}nchez-Carballido, Orti Senderos, Marcos Nieto, and Oihana Otaegui.
\newblock Semi-automatic cloud-native video annotation for autonomous driving.
\newblock \emph{Applied Sciences}, 10\penalty0 (12):\penalty0 4301, 2020.

\bibitem[Schott et~al.(2018)Schott, Rauber, Bethge, and
  Brendel]{ref4schott2018towards}
Lukas Schott, Jonas Rauber, Matthias Bethge, and Wieland Brendel.
\newblock Towards the first adversarially robust neural network model on mnist.
\newblock \emph{arXiv preprint arXiv:1805.09190}, 2018.

\bibitem[Shailesh and Judy()]{shailesh976automatic}
S~Shailesh and MV~Judy.
\newblock Automatic annotation of dance videos based on foot postures.
\newblock \emph{Indian Journal of Computer Science And Engineering, Engg
  Journals Publications-ISSN}, 976:\penalty0 5166.

\bibitem[Shankar et~al.(2020)Shankar, Roelofs, Mania, Fang, Recht, and
  Schmidt]{shankar2020evaluating}
Vaishaal Shankar, Rebecca Roelofs, Horia Mania, Alex Fang, Benjamin Recht, and
  Ludwig Schmidt.
\newblock Evaluating machine accuracy on imagenet.
\newblock In \emph{International Conference on Machine Learning}, pages
  8634--8644. PMLR, 2020.

\bibitem[Siddiqui et~al.(2015)Siddiqui, Mishra, and Verma]{siddiqui2015survey}
Adnan Siddiqui, Nischcol Mishra, and Jitendra~Singh Verma.
\newblock A survey on automatic image annotation and retrieval.
\newblock \emph{International Journal of Computer Applications}, 118\penalty0
  (20), 2015.

\bibitem[Song et~al.(2019)Song, Li, He, Li, Fang, and Chen]{song2019employing}
Yajun Song, Zeyuan Li, Jie He, Zesong Li, Xin Fang, and Dagang Chen.
\newblock Employing auto-annotated data for government document classification.
\newblock In \emph{Proceedings of the 2019 3rd International Conference on
  Innovation in Artificial Intelligence}, pages 121--125, 2019.

\bibitem[Song et~al.(2005)Song, Hua, Dai, and Wang]{song2005semi}
Yan Song, Xian-Sheng Hua, Li-Rong Dai, and Meng Wang.
\newblock Semi-automatic video annotation based on active learning with
  multiple complementary predictors.
\newblock In \emph{Proceedings of the 7th ACM SIGMM international workshop on
  Multimedia information retrieval}, pages 97--104, 2005.

\bibitem[Sorano et~al.(2020)Sorano, Carrara, Cintia, Falchi, and
  Pappalardo]{sorano2020automatic}
Danilo Sorano, Fabio Carrara, Paolo Cintia, Fabrizio Falchi, and Luca
  Pappalardo.
\newblock Automatic pass annotation from soccer videostreams based on object
  detection and lstm.
\newblock \emph{arXiv preprint arXiv:2007.06475}, 2020.

\bibitem[Sprengel et~al.(2016)Sprengel, Jaggi, Kilcher, and
  Hofmann]{sprengel2016audio}
Elias Sprengel, Martin Jaggi, Yannic Kilcher, and Thomas Hofmann.
\newblock Audio based bird species identification using deep learning
  techniques.
\newblock Technical report, 2016.

\bibitem[Stanchev et~al.(2020)Stanchev, Egbert, and Ruttenberg]{9031469}
Lubomir Stanchev, Hanson Egbert, and Benjamin Ruttenberg.
\newblock Automating deep-sea video annotation using machine learning.
\newblock In \emph{2020 IEEE 14th International Conference on Semantic
  Computing (ICSC)}, pages 17--24, 2020.
\newblock \doi{10.1109/ICSC.2020.00010}.

\bibitem[Sumathi et~al.(2011)Sumathi, Devasena, and
  Hemalatha]{sumathi2011overview}
T~Sumathi, C~Lakshmi Devasena, and M~Hemalatha.
\newblock An overview of automated image annotation approaches.
\newblock \emph{International Journal of Research and Reviews in Information
  Sciences}, 1\penalty0 (1):\penalty0 1--5, 2011.

\bibitem[Tabik et~al.(2017)Tabik, Peralta, Herrera-Poyatos, and
  Herrera]{ref2tabik2017snapshot}
Siham Tabik, Daniel Peralta, Andres Herrera-Poyatos, and Francisco Herrera.
\newblock A snapshot of image pre-processing for convolutional neural networks:
  case study of mnist.
\newblock \emph{International Journal of Computational Intelligence Systems},
  10\penalty0 (1):\penalty0 555--568, 2017.

\bibitem[Tay{\c{s}}i and Biricik()]{taycsimobile}
Ziya Tay{\c{s}}i and G{\"o}ksel Biricik.
\newblock A mobile ner annotation tool with turkish focus.

\bibitem[Torrey and Shavlik(2010)]{torrey2010transfer}
Lisa Torrey and Jude Shavlik.
\newblock Transfer learning.
\newblock In \emph{Handbook of research on machine learning applications and
  trends: algorithms, methods, and techniques}, pages 242--264. IGI global,
  2010.

\bibitem[Tousch et~al.(2012)Tousch, Herbin, and Audibert]{tousch2012semantic}
Anne-Marie Tousch, St{\'e}phane Herbin, and Jean-Yves Audibert.
\newblock Semantic hierarchies for image annotation: A survey.
\newblock \emph{Pattern Recognition}, 45\penalty0 (1):\penalty0 333--345, 2012.

\bibitem[Vondrick and Ramanan(2011)]{vondrick2011video}
Carl Vondrick and Deva Ramanan.
\newblock Video annotation and tracking with active learning.
\newblock \emph{Advances in Neural Information Processing Systems},
  24:\penalty0 28--36, 2011.

\bibitem[Wang(2011)]{wang2011survey}
Feichao Wang.
\newblock A survey on automatic image annotation and trends of the new age.
\newblock \emph{Procedia Engineering}, 23:\penalty0 434--438, 2011.

\bibitem[Wang et~al.(2017)Wang, Yan, Li, and Zhou]{wang2017employing}
Limin Wang, Qian Yan, Shoushan Li, and Guodong Zhou.
\newblock Employing auto-annotated data for person name recognition in judgment
  documents.
\newblock In \emph{Chinese Computational Linguistics and Natural Language
  Processing Based on Naturally Annotated Big Data}, pages 13--23. Springer,
  2017.

\bibitem[Wang and Hua(2011)]{wang2011active}
Meng Wang and Xian-Sheng Hua.
\newblock Active learning in multimedia annotation and retrieval: A survey.
\newblock \emph{ACM Transactions on Intelligent Systems and Technology (TIST)},
  2\penalty0 (2):\penalty0 1--21, 2011.

\bibitem[Wang et~al.(2006{\natexlab{a}})Wang, Hua, Dai, and Song]{4036892}
Meng Wang, Xian-sheng Hua, Li-rong Dai, and Yan Song.
\newblock Enhanced semi-supervised learning for automatic video annotation.
\newblock In \emph{2006 IEEE International Conference on Multimedia and Expo},
  pages 1485--1488, 2006{\natexlab{a}}.
\newblock \doi{10.1109/ICME.2006.262823}.

\bibitem[Wang et~al.(2006{\natexlab{b}})Wang, Hua, Song, Yuan, Li, and
  Zhang]{wang2006automatic}
Meng Wang, Xian-Sheng Hua, Yan Song, Xun Yuan, Shipeng Li, and Hong-Jiang
  Zhang.
\newblock Automatic video annotation by semi-supervised learning with kernel
  density estimation.
\newblock In \emph{Proceedings of the 14th ACM international conference on
  Multimedia}, pages 967--976, 2006{\natexlab{b}}.

\bibitem[Wang et~al.(2009)Wang, Hua, Hong, Tang, Qi, and Song]{wang2009unified}
Meng Wang, Xian-Sheng Hua, Richang Hong, Jinhui Tang, Guo-Jun Qi, and Yan Song.
\newblock Unified video annotation via multigraph learning.
\newblock \emph{IEEE Transactions on Circuits and Systems for Video
  Technology}, 19\penalty0 (5):\penalty0 733--746, 2009.

\bibitem[Xie et~al.(2020)Xie, Luong, Hovy, and Le]{xie2020self}
Qizhe Xie, Minh-Thang Luong, Eduard Hovy, and Quoc~V Le.
\newblock Self-training with noisy student improves imagenet classification.
\newblock In \emph{Proceedings of the IEEE/CVF Conference on Computer Vision
  and Pattern Recognition}, pages 10687--10698, 2020.

\bibitem[Xu et~al.(2019)Xu, Zhang, Wang, Wang, Chen, Gao, and
  Feng]{xu2019affective}
Jiahao Xu, Boyan Zhang, Zhiyong Wang, Yang Wang, Fang Chen, Junbin Gao, and
  David~Dagan Feng.
\newblock Affective audio annotation of public speeches with convolutional
  clustering neural network.
\newblock \emph{IEEE Transactions on Affective Computing}, 2019.

\bibitem[Xu et~al.(2011)Xu, Xue, and Zhou]{xu2011ensemble}
Xin-Shun Xu, Xiangyang Xue, and Zhi-Hua Zhou.
\newblock Ensemble multi-instance multi-label learning approach for video
  annotation task.
\newblock In \emph{Proceedings of the 19th ACM international conference on
  Multimedia}, pages 1153--1156, 2011.

\bibitem[Xu et~al.(2012)Xu, Jiang, Xue, and Zhou]{xu2012semi}
Xin-Shun Xu, Yuan Jiang, Xiangyang Xue, and Zhi-Hua Zhou.
\newblock Semi-supervised multi-instance multi-label learning for video
  annotation task.
\newblock In \emph{Proceedings of the 20th ACM international conference on
  Multimedia}, pages 737--740, 2012.

\bibitem[Yang et~al.(2018)Yang, Chen, Li, He, and Zhang]{yang2018distantly}
Yaosheng Yang, Wenliang Chen, Zhenghua Li, Zhengqiu He, and Min Zhang.
\newblock Distantly supervised ner with partial annotation learning and
  reinforcement learning.
\newblock In \emph{Proceedings of the 27th International Conference on
  Computational Linguistics}, pages 2159--2169, 2018.

\bibitem[Yimam et~al.(2014)Yimam, Biemann, de~Castilho, and
  Gurevych]{yimam2014automatic}
Seid~Muhie Yimam, Chris Biemann, Richard~Eckart de~Castilho, and Iryna
  Gurevych.
\newblock Automatic annotation suggestions and custom annotation layers in
  webanno.
\newblock In \emph{Proceedings of 52nd Annual Meeting of the Association for
  Computational Linguistics: System Demonstrations}, pages 91--96, 2014.

\bibitem[You et~al.(2018)You, Zhang, Hsieh, Demmel, and
  Keutzer]{you2018imagenet}
Yang You, Zhao Zhang, Cho-Jui Hsieh, James Demmel, and Kurt Keutzer.
\newblock Imagenet training in minutes.
\newblock In \emph{Proceedings of the 47th International Conference on Parallel
  Processing}, pages 1--10, 2018.

\bibitem[Zhang et~al.(2012)Zhang, Islam, and Lu]{zhang2012review}
Dengsheng Zhang, Md~Monirul Islam, and Guojun Lu.
\newblock A review on automatic image annotation techniques.
\newblock \emph{Pattern Recognition}, 45\penalty0 (1):\penalty0 346--362, 2012.

\bibitem[Zhang et~al.(2019)Zhang, Sun, Wang, Zhang, Yu, Zhang, Babyn, and
  Zhong]{zhang2019computer}
Shikun Zhang, Fengrong Sun, Naishun Wang, Cuicui Zhang, Qianlei Yu, Mingqiang
  Zhang, Paul Babyn, and Hai Zhong.
\newblock Computer-aided diagnosis (cad) of pulmonary nodule of thoracic ct
  image using transfer learning.
\newblock \emph{Journal of digital imaging}, 32\penalty0 (6):\penalty0
  995--1007, 2019.

\bibitem[Zheng et~al.(2019)Zheng, Zhang, Yusifov, and
  Shi]{zheng2019applications}
York Zheng, Qie Zhang, Anar Yusifov, and Yunzhi Shi.
\newblock Applications of supervised deep learning for seismic interpretation
  and inversion.
\newblock \emph{The Leading Edge}, 38\penalty0 (7):\penalty0 526--533, 2019.

\end{thebibliography}

\end{document}